\def\ARXIVVERSION{1}
\documentclass[11pt,letterpaper]{article}

\usepackage[margin=1in]{geometry}
\usepackage{amsmath,amssymb,amsfonts}
\usepackage{booktabs}
\usepackage{graphicx}
\usepackage{algorithm}
\usepackage{algpseudocode}
\usepackage{hyperref}
\usepackage{enumitem}
\usepackage{cite}

\title{Max-Q Selective Imitation for Human-in-the-Loop Online Robot Learning\\
with Monte Carlo Q-Chunk Critics}
\ifdefined\ARXIVVERSION
\author{Zihang Wang\thanks{Equal contribution.}
\and Yishan Wang\footnotemark[1]}
\else
\author{Anonymous Authors\\
\normalsize Paper-ID [anonymous]}
\fi
\date{}

\begin{document}
\maketitle

\begin{abstract}
Human-in-the-loop (HIL) online reinforcement learning for real robots must absorb human interventions quickly while continuing to improve beyond the human prior.
We present a training method for this setting based on two components.
First, an \emph{MC Q-chunk} critic regresses chunk-level action values onto Monte Carlo returns from the replay buffer, performing sample-average (behavior) policy evaluation so that intervention trajectories are credited directly rather than diluted by current-policy TD backups.
Second, \emph{max-Q selective imitation} updates the actor by imitating, at each state, the higher-$Q$ action between the current policy action and a buffer sample under a hard winner-take-all rule.
This rule automatically switches between learning from interventions and on-policy self-improvement: when the autonomous policy is stronger, targets align with the policy distribution, reducing the policy--target-sample gap that otherwise induces execution-time distribution shift.
In practice we score candidates with a standard critic ensemble mean to reduce comparison noise, without softening targets or introducing score-gap thresholds.
On a real USB pick-and-insertion task with 20 demonstrations, ACT QChunk-MCBC attains 99\% success within 30 minutes of HIL training, whereas HIL-SERL requires about 5 hours to converge.
In simulation on Peg Insertion and Square, ACT/Flow Q-chunk variants similarly reach $\ge$96\% success within roughly half an hour of effective training, outperforming HIL-SERL, EXPO, and E2HiL on the success--time frontier.
\end{abstract}

\section{Introduction}
\label{sec:intro}

Real-world robot reinforcement learning (RL) remains constrained by interaction cost, contact-rich failure modes, and limited exploration under safety requirements.
Human-in-the-loop (HIL) protocols address these constraints by combining autonomous rollouts with human interventions, yielding a replay buffer that mixes near-policy transitions with corrective actions that may lie far from the current policy $\pi$.
Effective learning in this regime requires both rapid uptake of useful human corrections and continued improvement beyond demonstration- or intervention-level performance.

Existing approaches only partially meet these requirements.
Behavioral cloning of interventions is stable but conservative.
Unconstrained maximization of $Q(s,a)$ over continuous actions invites out-of-distribution (OOD) overestimation.
Offline extractors such as Implicit Q-Learning (IQL)~\cite{kostrikov2022iql} keep updates in data support via advantage-weighted regression (AWR), yet AWR always imitates buffer actions and, under a mixed HIL buffer, repeatedly pulls $\pi$ across heterogeneous sample modes.
As a result, learning targets need not converge to the live policy even after $\pi$ improves, leaving a persistent gap between the \emph{policy distribution} and the \emph{target-sample distribution}.
Because closed-loop execution follows $\pi$, this gap exposes the robot to state--action regions underrepresented by the training targets and compounds tracking error.
SAC-style HIL systems such as HIL-SERL~\cite{luo2024hilserl} stay closer to the policy distribution, but their TD backups bootstrap under $a'\sim\pi$ rather than under the sample-average behavior in the buffer, which under-utilizes intervention returns.

\begin{figure*}[t]
\centering
\includegraphics[width=\textwidth]{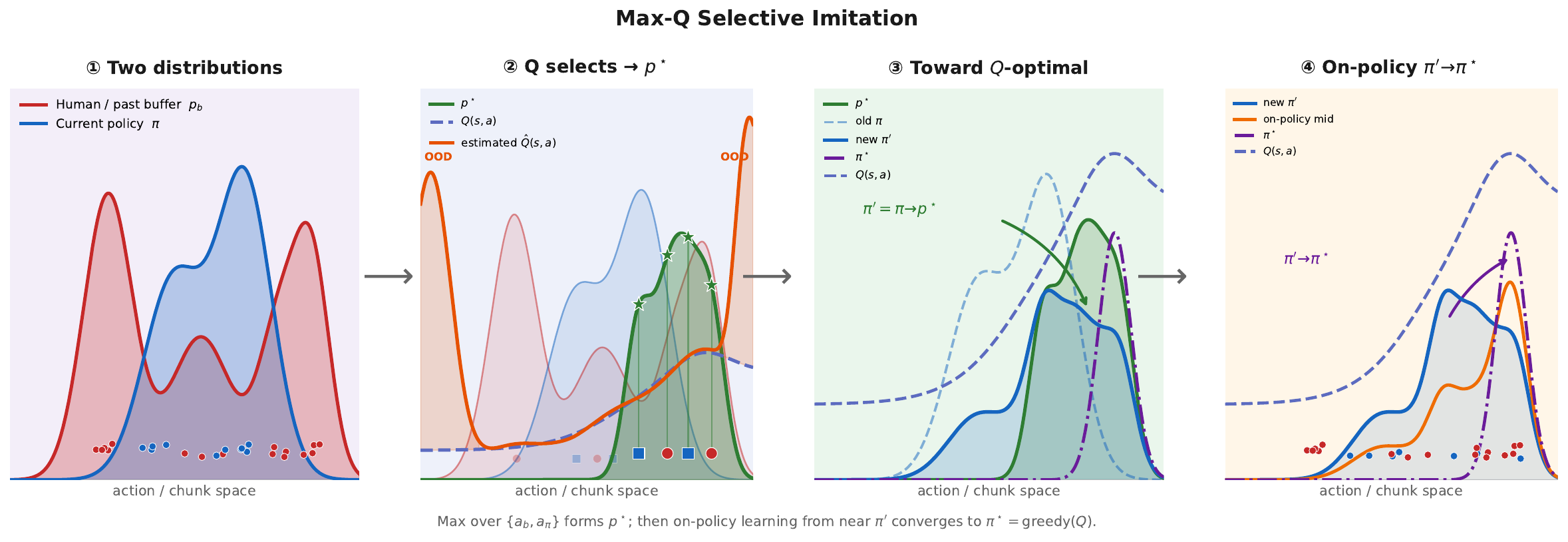}
\caption{Max-Q selective imitation in action / chunk space.
\textbf{(1)}~The replay buffer induces a human/past sample distribution $p_b$ while the learner maintains a current policy $\pi$.
\textbf{(2)}~An MC Q-chunk critic provides $Q(s,a)$; the estimated $\hat{Q}$ is reliable on support but overestimates out of distribution.
Hard $\max$ over candidates from $\{p_b,\pi\}$ selects winners that define a target $p^\star$.
\textbf{(3)}~The updated policy $\pi'$ lies between $\pi$ and $p^\star$ ($\pi'=\pi\!\rightarrow\! p^\star$).
\textbf{(4)}~On-policy learning near $\pi'$ further concentrates mass toward $\pi^\star=\mathrm{greedy}(Q)$.}
\label{fig:method}
\end{figure*}

We address these issues with a two-part method.
The critic learns chunk-level values by Monte Carlo regression onto logged returns (\emph{MC Q-chunk}), i.e., sample-average behavior evaluation that credits intervention paths.
The actor performs \emph{max-Q selective imitation}: for each state it compares $a_\pi=\pi(s)$ and a buffer action $a_b$, and regresses toward the hard winner
\begin{equation}
a^\star=\arg\max_{a\in\{a_\pi,a_b\}} Q(s,a).
\end{equation}
When interventions have higher $Q$, they are imitated; when $a_\pi$ wins, updates become on-policy and the target-sample distribution aligns with execution under $\pi$.
As an implementation detail, we evaluate $Q$ with a standard critic-ensemble mean to reduce early ranking noise while keeping the discrete, non-soft switching rule.
Figure~\ref{fig:method} summarizes the idea in action space.

\paragraph{Reading Figure~\ref{fig:method}.}
Panel~(1) shows the two sources that feed learning: the mixed human/past buffer distribution $p_b$ and the live policy $\pi$.
Panel~(2) scores candidates with $\hat{Q}$; because unconstrained maximization would chase OOD spikes of $\hat{Q}$, we only compare in-support actions and take a hard winner, so the selected samples prop up $p^\star$.
Panel~(3) makes the policy update explicit: $\pi'$ is an intermediate between the current $\pi$ and the Max-selected target $p^\star$, absorbing high-$Q$ human corrections without jumping fully onto the buffer.
Panel~(4) continues with on-policy improvement around $\pi'$, moving the mass toward the true optimum $\pi^\star$ implied by $Q$, closing the policy--target gap at execution time.

\paragraph{Contributions.}
\begin{itemize}[leftmargin=*,itemsep=1pt]
\item An MC Q-chunk critic for sample-average evaluation of mixed HIL trajectories, improving utilization of intervention returns relative to $\pi$-bootstrapped TD critics.
\item Max-Q selective imitation between policy and buffer actions, enabling label-free hard switching from intervention learning to on-policy self-improvement and reducing the policy--target-sample gap.
\item Simulation on Peg Insertion and Square, where ACT/Flow Q-chunk variants reach $\ge$96\% success within roughly half an hour of effective training and outperform HIL-SERL, EXPO, and E2HiL on the success--time frontier; plus real-robot USB pick-and-insertion, where ACT QChunk-MCBC reaches 99\% in 30 minutes while HIL-SERL converges in about 5 hours.
\end{itemize}

\section{Related Work}
\label{sec:related}

\paragraph{Human-in-the-loop robot RL.}
Interactive and HIL methods inject corrections or interventions during training~\cite{luo2024hilserl}.
HIL-SERL combines demonstrations, interventions, and off-policy SAC-style learning on hardware.
We share this protocol but replace $\pi$-conditioned TD evaluation with MC Q-chunk behavior evaluation, and replace likelihood-oriented actor updates with max-Q selective imitation.

\paragraph{Offline and offline-to-online RL.}
Offline RL limits OOD queries by constraining policies to data support~\cite{placeholder_offline,kostrikov2022iql}.
IQL evaluates values in-sample and extracts policies with AWR.
In online HIL fine-tuning, AWR continues to target buffer actions only, so improvement can cross-jump among modes rather than settle on the live policy.
Our actor explicitly includes $a_\pi$ as a candidate, allowing on-policy targets once $\pi$ dominates under $Q$.

\paragraph{Action chunking and Q-chunking.}
Action chunking is widely used for temporally coherent imitation.
Q-chunking~\cite{li2025qchunking} lifts TD actor--critic learning into a chunked action space.
We retain chunk-level $Q(s,\mathbf{a})$ but fit it with Monte Carlo returns so evaluation remains sample-average and does not bootstrap under the current $\pi$.

\paragraph{Max-over-candidates policy improvement.}
BCQ and EMaQ~\cite{placeholder_emaq} improve policies by maximizing $Q$ over actions proposed near behavior support.
Our variant is specialized to HIL: one candidate is always the live policy action and the other a replay sample (often an intervention), with improvement realized by supervised imitation of the hard winner.

\section{Method}
\label{sec:method}

\subsection{Problem Setup}
We consider an MDP $(\mathcal{S},\mathcal{A},P,r,\gamma)$ under a HIL online protocol.
The replay buffer $\mathcal{B}$ is initialized with a small demonstration set and expanded by autonomous rollouts together with human interventions.
Updates do not require intervention labels: each transition is treated as an ordinary state--action pair.
The objective is high task success under a limited robot interaction budget.

\subsection{MC Q-Chunk Critic}
Let $h$ be the chunk horizon and $\mathbf{a}_{t:t+h-1}=(a_t,\ldots,a_{t+h-1})$ an action chunk.
The critic $Q_\theta(s_t,\mathbf{a}_{t:t+h-1})$ scores the chunk from $s_t$.
Rather than TD bootstrapping under the current policy, we regress onto the Monte Carlo return along the logged trajectory,
\begin{equation}
G_t=\sum_{k=0}^{T_t-1}\gamma^k r_{t+k},
\end{equation}
and minimize
\begin{equation}
\mathcal{L}_Q=\mathbb{E}_{(s_t,\mathbf{a}_{t:t+h-1},G_t)\sim\mathcal{B}}\Big[\big(Q_\theta(s_t,\mathbf{a}_{t:t+h-1})-G_t\big)^2\Big].
\end{equation}
Because $G_t$ is generated by the buffer behavior---a mixture of autonomous and intervention actions---$Q_\theta$ estimates values under the sample-average behavior policy.
Intervention actions on informative trajectories raise MC targets directly and remain visible to subsequent max-Q selection.
Chunk-level scoring matches the temporal structure of demonstrations and interventions.

\paragraph{Implementation detail: critic ensemble.}
Hard switching is sensitive to early $Q$ noise.
As a standard variance-reduction trick~\cite{placeholder_redq}---not a methodological contribution---we optionally instantiate the ranking score as the mean of $M$ independently initialized critics trained on the same MC objective,
\begin{equation}
\bar{Q}(s,a)=\frac{1}{M}\sum_{i=1}^{M}Q_{\theta_i}(s,a),
\end{equation}
and write $Q\leftarrow\bar{Q}$ in the actor comparison below.
A single critic recovers the same learning rule.

\subsection{Max-Q Selective Imitation}
Pure buffer imitation cannot prefer an improving policy over stale logged actions.
Unconstrained $\arg\max_a Q(s,a)$ leaves the support where $Q$ is reliable.
We therefore compare two local candidates and imitate the hard winner.
For a sampled state $s$,
\begin{align}
a_\pi&=\pi_\phi(s), &
a_b&\sim\mathcal{B}(\cdot\mid s),\\
a^\star&=\arg\max_{a\in\{a_\pi,a_b\}} Q_\theta(s,a),
\end{align}
and
\begin{equation}
\mathcal{L}_\pi=\mathbb{E}_{s\sim\mathcal{B}}\big[\|\pi_\phi(s)-a^\star\|_2^2\big],
\end{equation}
optionally with an analogous chunk behavioral-cloning loss.
Stop-gradient through $a^\star$ and the $Q$ comparison prevents the actor objective from distorting the critic.
Importantly, the target remains a single discrete action: we do not mix $a_\pi$ and $a_b$ with soft weights, nor gate the switch by a hand-tuned score margin.
Soft targets would keep pulling $\pi$ toward buffer actions even after the policy is stronger, recreating the policy--target-sample gap that the hard rule is designed to close.

This construction has three operational consequences.
First, every scored action lies in the union of policy and data support, limiting OOD $Q$ queries.
Second, the hard $\max$ provides a directed improvement signal: the actor moves only toward the currently higher-$Q$ candidate.
Third, the same rule yields a label-free switch for HIL data.
High-$Q$ interventions are imitated when they outrank $a_\pi$; once $\pi$ is stronger, $a_\pi$ wins more often, so $\mathcal{L}_\pi$ regresses the policy onto itself and the target-sample distribution aligns with the policy (and thus with execution).
This reduces closed-loop exposure to regions underrepresented by training targets.
We use supervised regression onto $a^\star$ as the actor implementation; likelihood-weighted alternatives are left as optional baselines in experiments.

Online interaction still executes (optionally noisy) actions from $\pi$, preserving autonomous exploration whenever self-generated actions attain higher $Q$.

\subsection{Training Procedure}
Algorithm~\ref{alg:train} summarizes the loop (with the optional ensemble scoring trick).

\begin{algorithm}[t]
\caption{HIL Online Learning with MC Q-Chunk and Max-Q Imitation}
\label{alg:train}
\begin{algorithmic}[1]
\State Initialize $\mathcal{B}$ with demonstrations; initialize $Q_\theta$ (optionally an ensemble), $\pi_\phi$
\For{each online interaction step}
    \State Execute $a\sim\pi_\phi$ (with exploration); allow intervention
    \State Store transition/chunk in $\mathcal{B}$; compute $G_t$ on completed trajectories
    \State Sample minibatch from $\mathcal{B}$
    \State Update $Q_\theta$ by MC regression onto $G_t$
    \State Set $a^\star\leftarrow\arg\max_{a\in\{a_\pi,a_b\}}Q_\theta(s,a)$ and imitate $a^\star$
\EndFor
\end{algorithmic}
\end{algorithm}

\section{Experiments}
\label{sec:expts}

We evaluate HIL online learning in simulation on two contact-rich \texttt{gym\_hil} tasks---\emph{Peg Insertion} and \emph{Square}---against HIL-SERL~\cite{luo2024hilserl}, plain MCBC (no Q-chunk actor), EXPO, and E2HiL.
Our stack appears as \emph{ACT QChunk-MCBC} and \emph{Flow QChunk-MCBC}, which share the MC Q-chunk critic and max-Q selective imitation and differ only in the chunked action head.
Success is measured by autonomous evaluation episodes at saved checkpoints.
Curves are truncated at each method's best checkpoint (highest success, then lowest average episode length).
Time axes use logged pure training time; for all methods except HIL-SERL we zero the clock at the first checkpoint so early process warmup without interaction is excluded.
HIL-SERL retains its recorded optimization-window time.
When environment steps are not logged we impute them as $\beta\times$ training steps with $\beta\!\approx\!2.0$ fit on EXPO/E2HiL.

\subsection{Peg Insertion}
\label{sec:expts-peg}

\begin{figure}[t]
\centering
\begin{minipage}{0.49\linewidth}
\centering
\includegraphics[width=\linewidth]{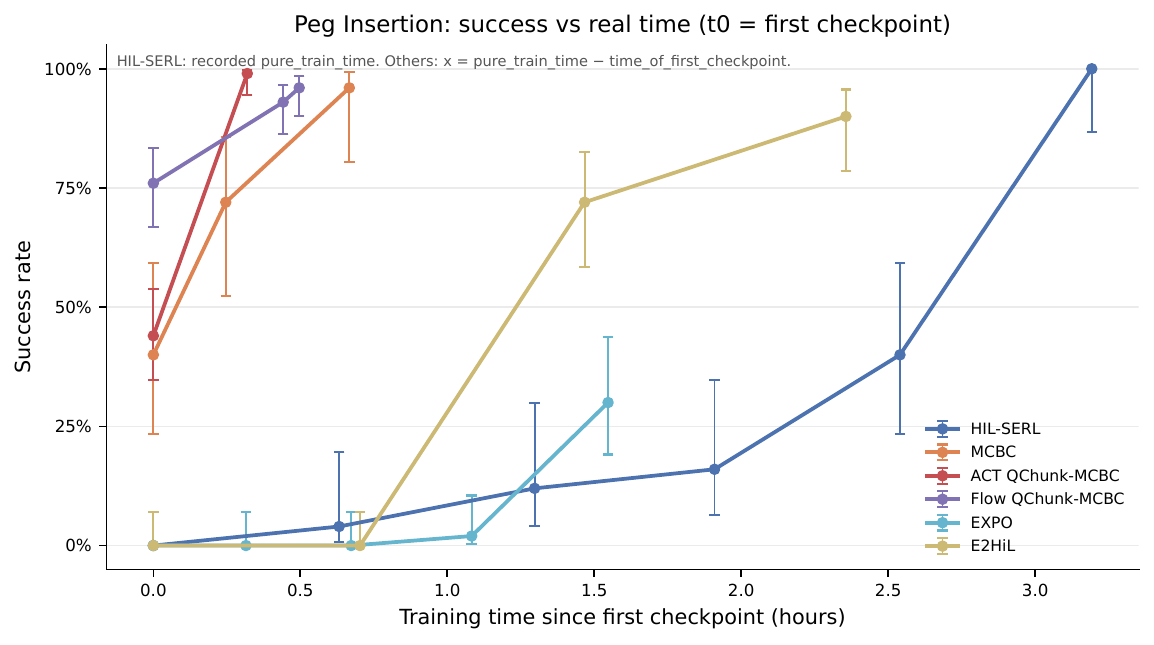}
\end{minipage}\hfill
\begin{minipage}{0.49\linewidth}
\centering
\includegraphics[width=\linewidth]{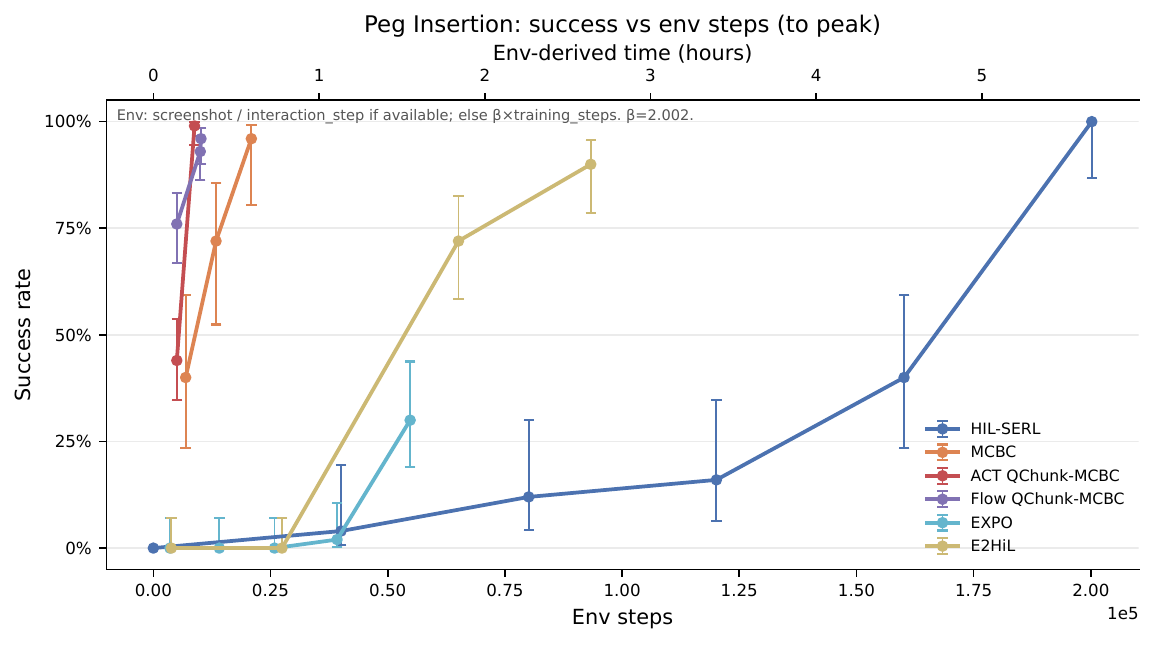}
\end{minipage}
\caption{Peg Insertion: success versus real training time (left; HIL-SERL uses recorded time, others start at the first checkpoint) and versus environment steps (right). Curves stop at each method's peak checkpoint.}
\label{fig:peg-curves}
\end{figure}

\begin{figure}[t]
\centering
\includegraphics[width=0.72\linewidth]{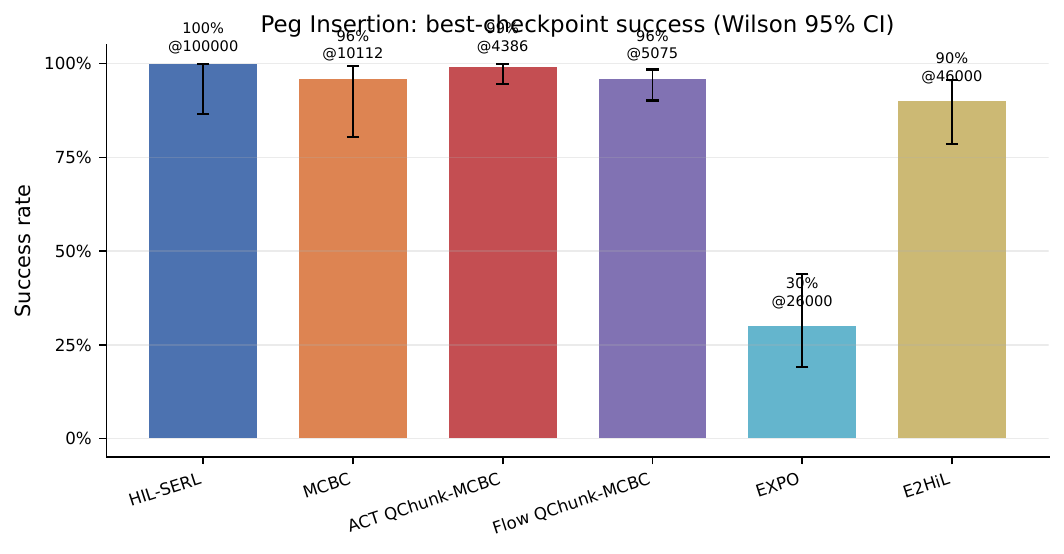}
\caption{Peg Insertion: best-checkpoint success with Wilson 95\% confidence intervals.}
\label{fig:peg-best}
\end{figure}

\begin{table}[t]
\caption{Peg Insertion: peak checkpoint summary}
\label{tab:peg}
\centering
\scriptsize
\begin{tabular}{lrrrr}
\toprule
Method & Steps & Env & Hours$^\dagger$ & Success \\
\midrule
HIL-SERL & 100\,000 & 200\,196 & 3.19 & 100\% \\
MCBC & 10\,112 & 20\,906 & 0.67 & 96\% \\
ACT QChunk-MCBC & 4\,386 & 8\,781 & 0.32 & 99\% \\
Flow QChunk-MCBC & 5\,075 & 10\,160 & 0.50 & 96\% \\
EXPO & 26\,000 & 54\,782 & 1.55 & 30\% \\
E2HiL & 46\,000 & 93\,315 & 2.36 & 90\% \\
\bottomrule
\end{tabular}\\[0.35em]
{\footnotesize $^\dagger$Hours after first checkpoint (HIL-SERL: recorded pure train time).}
\end{table}

Figure~\ref{fig:peg-curves} and Table~\ref{tab:peg} summarize Peg Insertion.
Both Q-chunk variants reach near-ceiling success much earlier than HIL-SERL and E2HiL: ACT QChunk-MCBC attains 99\% by $0.32$\,h after the first checkpoint ($\approx$4.4k training steps), and Flow QChunk-MCBC reaches 96\% by $0.50$\,h.
Plain MCBC also climbs to 96\% by $0.67$\,h, showing that selective imitation of high-$Q$ buffer actions already helps, while action chunking further shortens the time to peak under the same MC critic.
HIL-SERL eventually matches 100\% success but only after $\approx$3.2\,h / 100k steps, reflecting slower uptake of interventions when TD backups remain on-policy.
E2HiL peaks at 90\% near 2.4\,h; EXPO saturates at 30\% and does not recover within the logged window (Figure~\ref{fig:peg-best}).
Overall, the MC Q-chunk + max-Q stack delivers the best success--time trade-off on this task.

\subsection{Square}
\label{sec:expts-square}

\begin{figure}[t]
\centering
\begin{minipage}{0.49\linewidth}
\centering
\includegraphics[width=\linewidth]{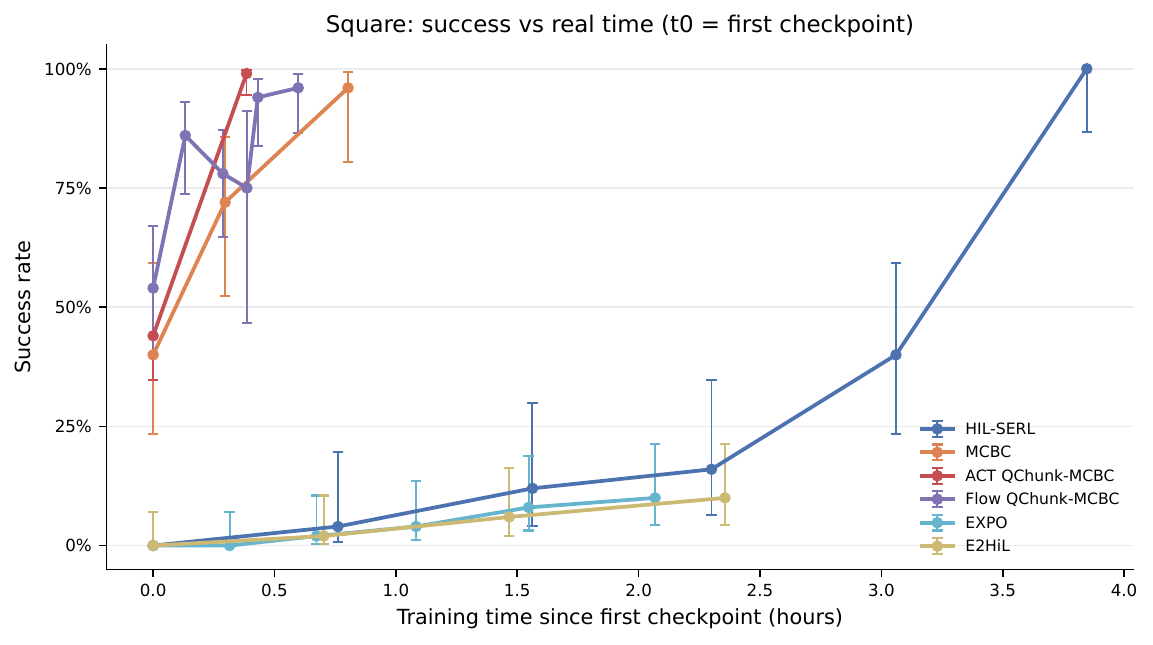}
\end{minipage}\hfill
\begin{minipage}{0.49\linewidth}
\centering
\includegraphics[width=\linewidth]{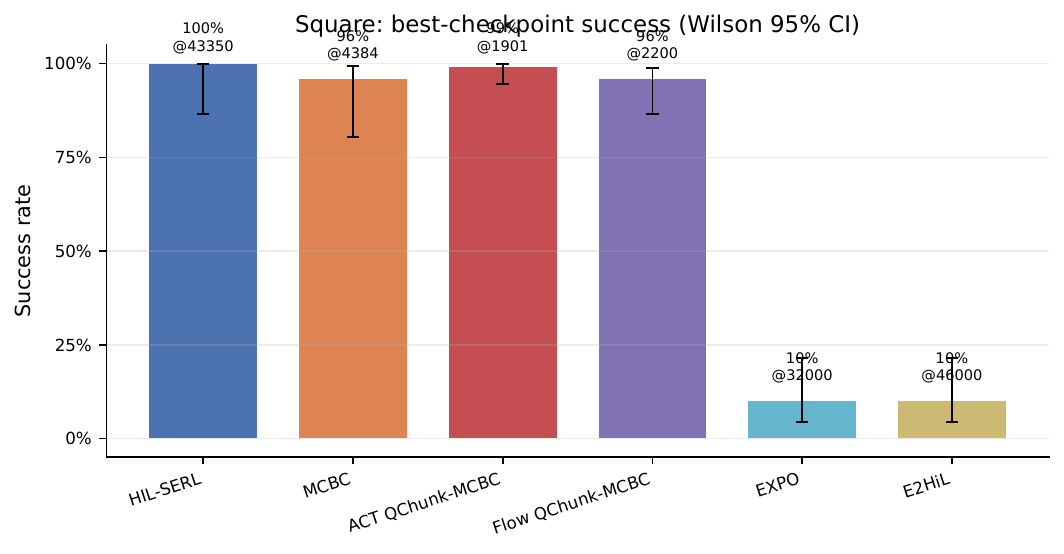}
\end{minipage}
\caption{Square: success versus training time since the first checkpoint (left) and best-checkpoint success (right).}
\label{fig:square-curves}
\end{figure}

\begin{table}[t]
\caption{Square: peak checkpoint summary}
\label{tab:square}
\centering
\scriptsize
\begin{tabular}{lrrrr}
\toprule
Method & Steps & Env & Hours$^\dagger$ & Success \\
\midrule
HIL-SERL & 43\,350 & 339\,626 & 3.85 & 100\% \\
MCBC & 4\,384 & 35\,466 & 0.80 & 96\% \\
ACT QChunk-MCBC & 1\,901 & 14\,896 & 0.38 & 99\% \\
Flow QChunk-MCBC & 2\,200 & 17\,236 & 0.60 & 96\% \\
EXPO & 32\,000 & 72\,031 & 2.07 & 10\% \\
E2HiL & 46\,000 & 93\,315 & 2.36 & 10\% \\
\bottomrule
\end{tabular}\\[0.35em]
{\footnotesize $^\dagger$Same time convention as Table~\ref{tab:peg}.}
\end{table}

On Square (Figure~\ref{fig:square-curves}, Table~\ref{tab:square}), the same ranking appears.
Flow QChunk-MCBC rises from mid-training success to a 96\% peak at 2\,200 steps / $0.60$\,h after the first checkpoint.
ACT QChunk-MCBC again reaches the highest peak earliest (99\% by $0.38$\,h), while plain MCBC hits 96\% by $0.80$\,h.
HIL-SERL eventually reaches 100\% but only after several hours of training.
EXPO and E2HiL remain near failure, peaking at 10\%.
Thus Square confirms that the MC Q-chunk + max-Q stack is sample- and time-efficient on a second contact-rich assembly task, whereas competing HIL baselines do not close the gap under the same protocol.

\subsection{Real-robot USB insertion}
\label{sec:expts-real}

\begin{figure}[t]
\centering
\begin{minipage}{0.49\linewidth}
\centering
\includegraphics[width=\linewidth]{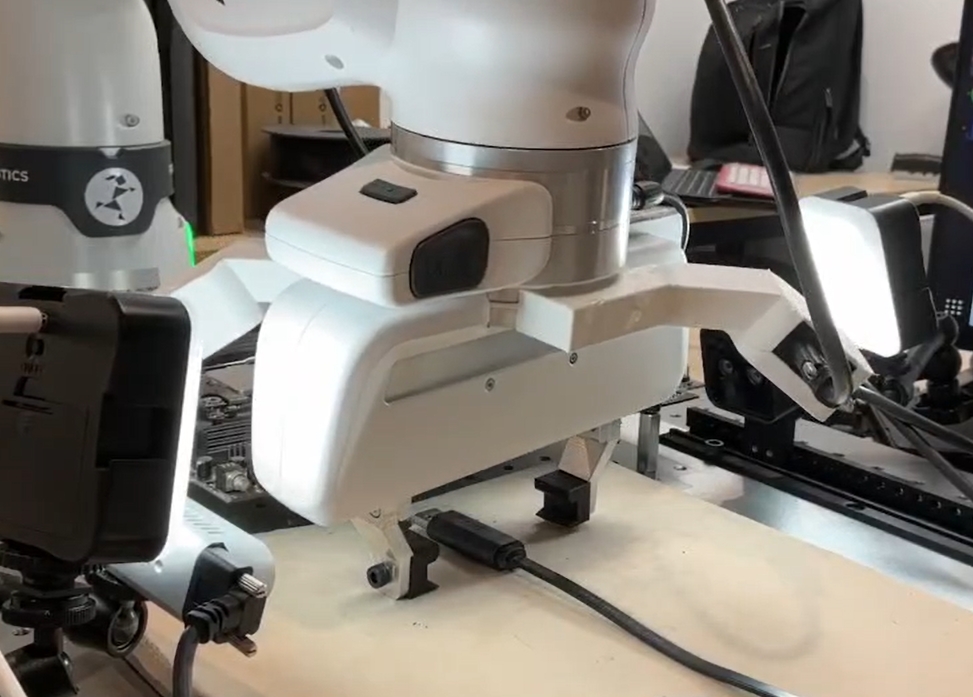}
\end{minipage}\hfill
\begin{minipage}{0.49\linewidth}
\centering
\includegraphics[width=\linewidth]{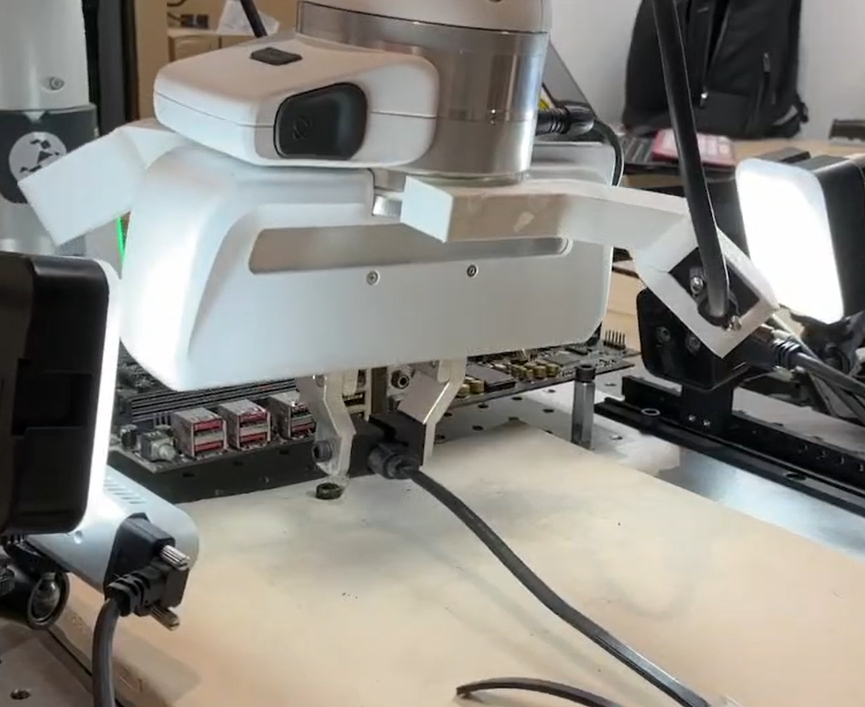}
\end{minipage}
\caption{Real-robot USB pick-and-insertion setup. Left: workspace with dual LED illumination and a Franka-class arm holding a USB connector. Right: close-up of alignment into a motherboard rear I/O port.}
\label{fig:usb-setup}
\end{figure}

\begin{figure}[t]
\centering
\begin{minipage}{0.55\linewidth}
\centering
\includegraphics[width=\linewidth]{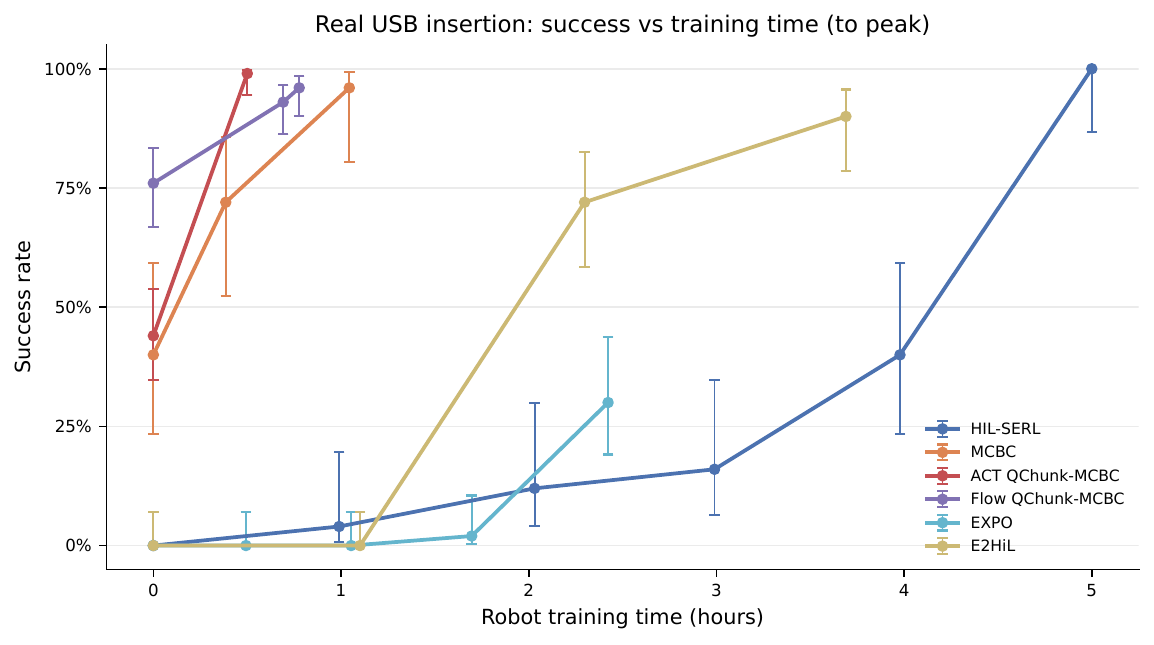}
\end{minipage}\hfill
\begin{minipage}{0.42\linewidth}
\centering
\includegraphics[width=\linewidth]{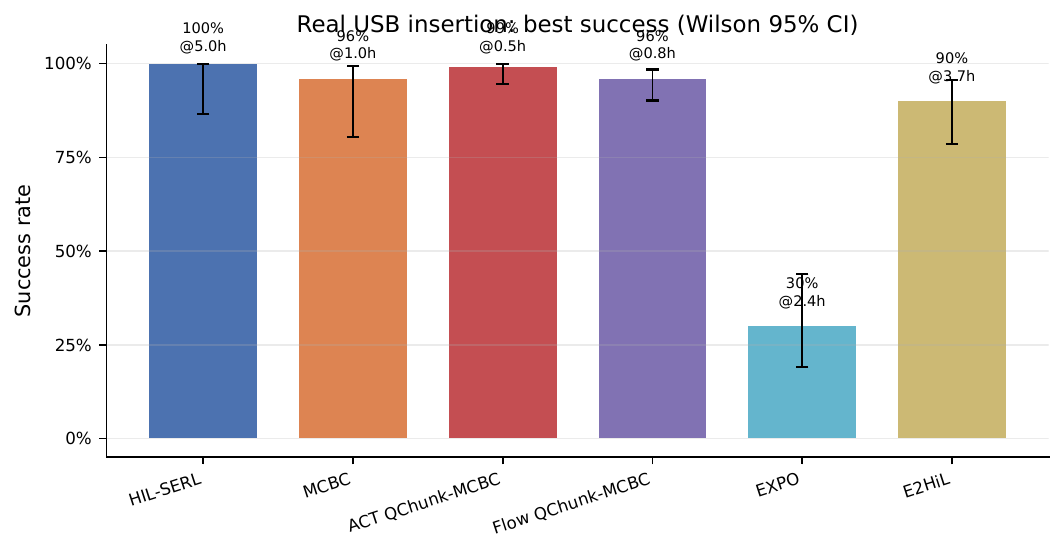}
\end{minipage}
\caption{Real USB insertion: success versus robot training time (left; curves truncated at each method's peak) and best-checkpoint success (right).}
\label{fig:usb-curves}
\end{figure}

\begin{table}[t]
\caption{Real USB insertion: peak training time and success}
\label{tab:main}
\centering
\scriptsize
\begin{tabular}{lrr}
\toprule
Method & Time to peak & Success \\
\midrule
HIL-SERL & 5.0\,h & 100\% \\
MCBC & 1.0\,h & 96\% \\
ACT QChunk-MCBC & 0.5\,h (30\,min) & 99\% \\
Flow QChunk-MCBC & 0.8\,h & 96\% \\
EXPO & 2.4\,h & 30\% \\
E2HiL & 3.7\,h & 90\% \\
\bottomrule
\end{tabular}
\end{table}

We evaluate on a real USB pick-and-insertion task (Figure~\ref{fig:usb-setup}): a Franka-class arm with a wrist camera and parallel gripper must grasp a USB plug and insert it into a motherboard rear I/O port under dual LED lighting.
Training uses 20 human demonstrations, autonomous rollouts, and human interventions.
Figure~\ref{fig:usb-curves} and Table~\ref{tab:main} report success against robot training time.
HIL-SERL reaches ceiling success after approximately 5 hours of HIL training.
Under the same protocol, ACT QChunk-MCBC attains 99\% by 30 minutes, Flow QChunk-MCBC reaches 96\% by roughly 0.8 hours, and plain MCBC reaches 96\% by about 1 hour---an order-of-magnitude reduction relative to HIL-SERL.
E2HiL peaks near 90\% only after several hours, and EXPO saturates at 30\%.
These real-robot trends mirror the simulation success--time frontier: MC Q-chunk critics with max-Q selective imitation absorb interventions quickly and converge far sooner than TD-based HIL baselines.

\subsection{Ablations}
\label{sec:expts-ablations}
To isolate the two components, we recommend (and plan to report) the following comparisons:
(i) MC Q-chunk versus $\pi$-bootstrapped TD critics;
(ii) actor variants including full buffer imitation, SAC-style actors, unconstrained $Q$-maximization, online IQL/AWR, and max-Q selective imitation;
(iii) the fraction of updates in which $a_\pi$ wins over $a_b$ over training time;
(iv) win rates stratified by intervention versus autonomous buffer actions (labels for analysis only);
(v) sensitivity to demonstration count and intervention frequency;
(vi) as an optional check on the ensemble scoring trick, single critic versus ensemble mean.

\section{Limitations}
\label{sec:limitations}

The switching decision depends on critic quality; noisy early $Q$ estimates can select suboptimal targets (partially mitigated in practice by a critic ensemble, which remains a heuristic rather than a complete fix).
Monte Carlo returns typically have higher variance than TD backups and may require sufficient episode completion for stable targets.
The chunk horizon $h$ remains a task-dependent hyperparameter.
Broader multi-task evaluation and a formal characterization of the two-candidate improvement operator are left for future work.

\section{Conclusion}
\label{sec:conclusion}

We presented a HIL online robot learning method that combines an MC Q-chunk critic for sample-average behavior evaluation with hard max-Q selective imitation between policy and buffer actions.
The resulting updates absorb useful interventions when they improve $Q$, and shift toward on-policy self-improvement as the autonomous policy strengthens, reducing the policy--target-sample gap at execution time.
In simulation on Peg Insertion and Square, ACT/Flow Q-chunk variants reach $\ge$96\% success within roughly half an hour of effective training and dominate HIL-SERL, EXPO, and E2HiL on the success--time frontier; on real USB pick-and-insertion, ACT QChunk-MCBC reaches 99\% in 30 minutes while HIL-SERL requires about 5 hours to converge.

\ifdefined\ARXIVVERSION
\else
\section*{Acknowledgments}
Omitted for anonymous review.
\fi

\end{document}